\documentclass[runningheads, envcountsame, a4paper]{llncs}
\usepackage{graphicx}
\usepackage{amsmath,amsfonts,bm}

\def\eqref#1{equation~\ref{#1}}
\def\1{\bm{1}}

\DeclareMathAlphabet{\mathsfit}{\encodingdefault}{\sfdefault}{m}{sl}
\SetMathAlphabet{\mathsfit}{bold}{\encodingdefault}{\sfdefault}{bx}{n}

\DeclareMathOperator*{\argmin}{arg\,min}

\usepackage{cleveref}
\usepackage{algorithm, algorithmic}
\usepackage{caption}
\usepackage{wrapfig}
\usepackage{dirtytalk}
\usepackage{comment}
\usepackage{booktabs}
\usepackage[misc]{ifsym}
\usepackage{array}
\usepackage{url}
\usepackage{empheq}
\usepackage{xcolor}

\usepackage[algo2e,ruled,vlined]{algorithm2e}
\newcolumntype{P}[1]{>{\centering\arraybackslash}p{#1}}

\newcounter{equationset}
\begin{document}

\title{Differentiable Feature Selection, a Reparameterization Approach}
\toctitle{Differentiable Feature Selection, a Reparameterization Approach}


\author{J\'{e}r\'{e}mie Don\`{a} \textsuperscript{\Letter, } \inst{1} \and Patrick Gallinari\inst{1,2}}
\tocauthor{J\'{e}r\'{e}mie~Don\`{a}}
\tocauthor{Patrick Gallinari}

%Second Author\inst{2,3}\orcidID{1111-2222-3333-4444} \and
%Third Author\inst{3}\orcidID{2222--3333-4444-5555}}

%
\authorrunning{J. Don\`{a} and P. Gallinari }
\institute{
Sorbonne Université, CNRS, LIP6, F-75005 Paris, France
\and
Criteo AI Labs, Paris, France \\
\email{firstname.lastname@lip6.fr}}

\maketitle 

\begin{abstract}
We consider the task of feature selection for reconstruction which consists in choosing a small subset of features from which whole data instances can be reconstructed.
This is of particular importance in several contexts involving for example costly physical measurements, sensor placement or information compression.
To break the intrinsic combinatorial nature of this problem, we formulate the task as optimizing a binary mask distribution enabling an accurate reconstruction. 
We then face two main challenges. 
One concerns differentiability issues due to the binary distribution. 
The second one corresponds to the elimination of redundant information by selecting variables in a correlated fashion which requires modeling the covariance of the binary distribution.
We address both issues by introducing a relaxation of the problem via a novel reparameterization of the logitNormal distribution. 
We demonstrate that the proposed method provides an effective exploration scheme and leads to efficient feature selection for reconstruction through evaluation on several high dimensional image benchmarks.
We show that the method leverages the intrinsic geometry of the data, facilitating reconstruction.
\end{abstract}


\section{Introduction}
Learning sparse representations of data finds essential real-world applications as in budget learning where the problem is limited by the number of features available or in embedded systems where the hardware imposes computational limitations. 
Feature selection serves similar objectives giving insights about variable dependencies and reducing over-fitting \cite{guyon_introduction_2003}.
Combined with a reconstruction objective, feature selection is a sensible problem when collecting data is expensive which is often the case with physical processes.
For example, consider optimal sensor placement.
This task consists in optimizing the location of sensors measuring a scalar field over an area of interest (e.g pressure, temperature) to enable truthful reconstruction of the signal on the whole area.
It finds applications in climate science \cite{haeberli_integrated_2007,mcphaden_rama:_2009}, where key locations are monitored to evaluate the impact of climate change on snow melt and Monsoon. These examples illustrate how feature selection for reconstruction may be critically enabling for large scale problems where measurements are costly.

Common practices for feature selection involves a $\ell_1$-regularization over the parameters of a linear model to promote sparsity \cite{tibshirani_regression_1996}.
Initiated by \cite{tibshirani_regression_1996}, several refinements have been developed for feature selection.
For example, \cite{yang_unsupervised_2019} employs a $\ell_{2,1}$-norm in a linear auto-encoder.
\cite{han_autoencoder_2017} impose a $\ell_1$-penalty on the first layer of a deep auto-encoder to select features from the original signal.
Finally, Group-Lasso methods extended lasso by applying the sparse $\ell_1$-penalty over precomputed chunks of variables to take prior knowledge into account while selecting features.
Theses approaches suffer from two main limitations: the design of the groups for Group-Lasso methods and the loss of the intrinsic structure of the data as
both \cite{yang_unsupervised_2019,han_autoencoder_2017} treat the input signal as a vector.
Moreover, non-linear $\ell_1$ based methods for feature selection and reconstruction are intrinsically ill posed, see \cref{sup.l1}.
Like Group-Lasso methods, our proposition aims at selecting variables in a correlated fashion, to eliminate redundant information, while leveraging the structure of the data.
We illustrate its efficiency on images but it can be adapted to exploit patterns in other types of structured data as graphs.

We propose a novel sparse embedding method that can tackle feature selection through an end-to-end-approach.
To do so, we investigate the learning of binary masks sampled from a distribution over binary matrices of the size of the image, with $1$ indicating a selected pixel.
We alleviate differentiability issues of learning categorical variables by relying on a continuous relaxation of the problem.
The learned latent binary distribution is optimized via a stochastic exploration scheme.
We consider the dependency between the selected pixels and we propose to sample the pixels in the mask in a correlated fashion to perform feature selection efficiently.
Accordingly, we learn a correlated logitNormal distribution via the reparameterization trick allowing for an efficient exploration of the masks space while preserving structural information for reconstruction.
Finally, sparsity in the embedding is enforced via a relaxation of the $\ell_0$-norm.
To summarize, we aim at learning a binary mask for selecting pixels from a distribution of input signals $x$, with $x \in\mathbb{R}^{n \times n}$ for images, enabling an accurate reconstruction.
We formulate our problem as learning jointly a parametric sampling operator $S$ which takes as input a random variable $z\in\mathcal{Z}\subseteq \mathbb{R}^d$ and outputs binary masks, i.e. $S: \mathcal{Z} \rightarrow \{0, 1\}^{n\times n}$. 
We introduce two ways to learn the sampling operator $S$.
For reconstruction, an additional operator denoted $G$ learns to reconstruct the data $x$ from the sparse measurements $s \odot x$.
Our proposed approach is fully differentiable and can be optimized directly via back-propagation. Our main contributions are:
\begin{itemize}
\item  We introduce a correlated logitNormal law to learn sparse binary masks, optimized thanks to the reparameterization trick. 
This reparameterization is motivated statistically. Sparsity is enforced via a relaxed $\ell_0$-norm.
\item We formulate the feature selection task for 2-D data as the joint learning of a binary mask and a reconstruction operator and propose a novel approach to learn the parameters of the considered logitNormal law.
\item We evidence the efficiency of our approach on several datasets: Mnist, CelebA and a complex geophysical dataset.
\end{itemize}
\section{Related Work}
Our objective of learning binary mask lies in between a few major domains: density modeling, feature selection and compressed sensing.

\subsubsection{Density Modeling via Reparameterization}
Sampling being not differentiable, different solutions have been developed in order to estimate the gradients of the parameters of a sampling operator.
Gradient estimates through score functions \cite{williams_simple_1992,bengio_estimating_2013} usually suffer from high variance or bias. Reparameterization \cite{kingma_auto-encoding_2013} provides an elegant way to solve the problem. 
It consists in sampling from a fixed distribution serving as input to a parametric transformation in order to obtain both the desired distribution and the gradient with respect to the parameters of interest.
However, the learning of categorical variables remains tricky as optimizing on a discrete set lacks differentiability.
Continuous relaxation of discrete variables enables parameters optimization through the reparameterization trick.
Exploiting this idea, \cite{maddison_concrete_2016,Jang17} developed the concrete law as a reparameterization of the Gumbel max variable for sampling categorical variables \cite{luce_individual_1959}.
Alternative distributions, defining relaxations of categorical variables can be learned by reparameterization such as the Dirichlet or logitNormal distribution \cite{Figurnov2018,kocisky_semantic_2016}.
Nonetheless, most previous approaches learn factorized distribution, thus selecting variables independently when applied to a feature selection task.
In contrast, we rely on the logitNormal distribution to propose a reparameterization scheme enabling us to sample the mask pixels jointly, taking into account dependencies between them and exploiting the patterns present in 2-D data.

\subsubsection{Feature Selection} \label{related_2}
Wrapper methods, \cite{guyon_introduction_2003,xing_feature_2001,maldonado_wrapper_2009} select features for a downstream task whereas filter methods \cite{he_laplacian_2006,yu_feature_2003,koller_toward_1996} rank the features according to tailored statistics. Our work belongs to the category of \textit{embedded} methods, that address selection as part of the modeling process. 
$\ell_1$-penalization over parameters, as for instance in Lasso and in Group Lasso variants \cite{Yuan2006,simon_sparse-group_2013,zhou_exclusive_2010}, is a prototypical embedded method.
$\ell_1$-penalty was used for feature selection for example in \cite{zhu_unsupervised_2015,yang_unsupervised_2019} learning a linear encoding with a $\ell_{2,1}$-constraint for a reconstruction objective. 
Auto-encoders \cite{Hinton2006} robustness to noise and sparsity is also exploited for feature selection  \cite{vincent_extracting_2008,makhzani_k-sparse_2013,ng_feature_2004}.
For example, AEFS \cite{han_autoencoder_2017} extends Lasso with non linear auto-encoders, generalizing \cite{zhu_unsupervised_2015}.
Another line of work learns embeddings preserving local properties of the data and then find the best variables in the original space to explain the learned embedding, using either $\ell_1$ or $\ell_{2,1}$ constraints \cite{deng_unsupervised_nodate,hou_joint_2014}. 
Closer to our work, \cite{abid_concrete_2019} learn a matrix of weights $m$, where each row follow a concrete distribution \cite{maddison_concrete_2016}. 
That way each row of matrix $m$ samples one feature in x.
The obtained linear projection $m.x$ is decoded by a neural network, and $m$ is trained to minimize the $\ell_2$-loss between reconstructions and targets. 
Because $x$ is treated as a vector, here too, the structure of the data is ignored and lost in the encoding process.
%These methods do not consider the data intrinsic structure such as 2-D patterns in images.
Compared to these works, we leverage the dependencies between variables in the 2-D pixel distribution, by sampling binary masks via an adaptation of the logitNormal distribution.

\subsubsection{Compressed Sensing} \label{sec.cs}
Our work is also related to compressed sensing (CS) where the objective is to reconstruct a signal from limited (linear) measurements \cite{donoho_compressed_2006}.
Deep learning based compressed sensing algorithms have been developed recently: \cite{pmlr-v70-bora17a} use a pre-trained generative model and optimize the latent code to match generated measurements to the true ones;
The measurement process can be optimized along with the reconstruction network as in \cite{wu_deep_2019}.
Finally, \cite{manohar_data-driven_2018} use a CS inspired method based on the pivots of a QR decomposition over the principal components matrix to optimize the placement of sensors for reconstruction, but scales poorly for large datasets.
Our approach differs from CS.
Indeed, for CS, measurements are linear combinations of the signal sources, whereas we consider pixels from the original image.
Thus, when CS aims at reconstructing from linear measurements, our goal is to preserve the data structural information to select a minimum number of variables for reconstruction.
\section{Method}
We now detail our framework to learn correlated sparse binary masks for feature selection and reconstruction through an end-to-end approach.
The choice of the logitNormal distribution, instead of the concrete distribution \cite{maddison_concrete_2016}, is motivated by the simplicity to obtain correlated variables thanks to the stability of independent Gaussian law by addition as detailed below.
We experimentally show in \cref{sec.exp} that taking into account such correlations helps the feature selection task.
This section is organised as follows : we first introduce in \cref{subsec:preli} some properties of the logitNormal distribution and sampling method for this distribution. 
We detail in \cref{subsec:param} our parameterization for the learning of the masks distribution.
Finally, in \cref{sec.mask_sparsity} we show how to enforce sparsity in our learned distribution before detailing our reconstruction objective in \cref{subsec:reconstruction}.

\subsection{Preliminaries: logitNormal Law on ${[0, 1]}$ } \label{subsec:preli}
Our goal is to sample a categorical variable in a differentiable way. 
We propose to parameterize the sampling on the simplex by the logitNormal law, introduced in \cite{aitchison_logistic-normal_1980}. 
We detail this reparameterization scheme for the unidimensional case since we aim at learning binary encodings.
It can be generalized to learn k-dimensional one-hot vector, see supplementary materials \cref{sup.ln_muli}.
Let ${z \sim \mathcal{N}(\mu, \sigma)}$, and $Y$ defined as:
\begin{equation}
 Y = \text{sigmoid}(z) \label{eq.logitN_one_dim}
\end{equation}
Then $Y$ is said to follow a logitNormal law. This distribution defines a probability over $[0,1]$, admits a density and its cumulative distribution function has an analytical expression used to enforce sparsity in \cref{sec.mask_sparsity}.

\begin{figure}[h!]
\centering
 \includegraphics[width=0.72\textwidth,trim=0 0 0 0.61cm, clip]{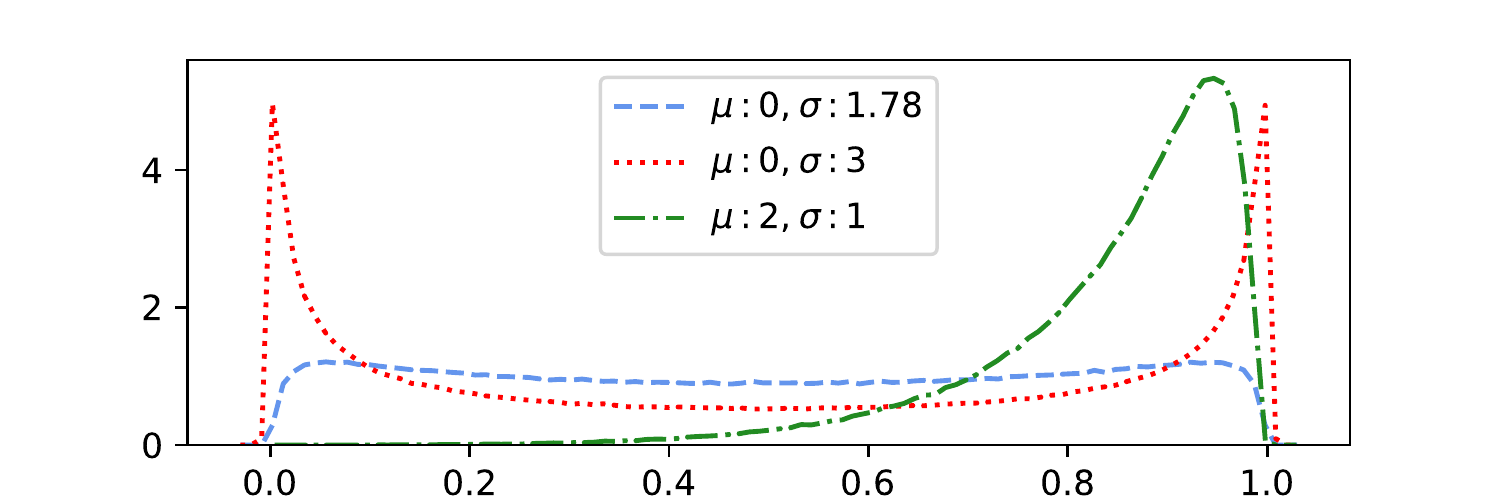}
 \vspace{-0.1in}
 \caption{Density of the logitNormal law for various couple $(\mu,\sigma)$: $(\mu=0,\sigma=1.78)$ (dashed line), $(\mu=0, \sigma=3)$ (dotted line) $(\mu=2, \sigma=1)$ (dotted and dashed). 
 %The dashed line ($\sigma=1.78$) is flat and relatively \textit{uniform} the other are either mono bi modal
 }
 \label{fig:std}
\end{figure}

This distribution can take various forms as shown in \cref{fig:std} and be flat as well as bi-modal.
By introducing a temperature in the sigmoid so that we have, $\text{sigmoid}_\lambda(z) = \frac{1}{1+\exp^{-z/\lambda}}$, we can polarize the logitNormal distribution.
In \Cref{prop1} we evidence the link between the 0-temperature logitNormal distribution and Bernoulli distribution:

\begin{proposition}[Limit Distribution] \label{prop1}
Let $W \in \mathbb{R}^n$ be a vector and $b \in \mathbb{R}$ a scalar.
Let
${Y=\text{sigmoid}_\lambda(W.z^T+b)}$ \label{eq.reparam}, where ${z \sim \mathcal{N}(0, I_n)}$, when $\lambda$ decrease towards $0$, $Y$ converges in law towards a Bernoulli distribution and we have:
\begin{align}
 &\lim_{\lambda \rightarrow 0} \mathbb{P}(Y=1) = 1 - \Phi\big (\frac{-b}{\sqrt{\sum_i w_i ^2}}\big) 
\label{eq.lim_1} \\
 &\lim_{\lambda \rightarrow 0} \mathbb{P}(Y=0) = \Phi\big(\frac{-b}{\sqrt{\sum_i w_i ^2}}\big) \label{eq.lim_0}
\end{align}
Where $\Phi$ is the cumulative distribution function of the Normal law $\mathcal{N}(0,1)$,
\end{proposition}
The proof is available in supplementary, \cref{proof.prop1}.
\Cref{eq.reparam} characterizes the limit distribution as the temperature goes down to $0$, and $Y$ defines a differentiable relaxation of a Bernoulli variable. 
This proposition is used to remove randomness in our learned mask distribution, see \cref{sec.exp}.

We relax the objective of learning of a binary mask in $\{0, 1\}$ by learning in $[0,1]$ using the logitNormal law.
Let $m\in\mathbb{N}$, be the dimension of the desired logitNormal variable $Y$.
A simple solution for learning the logitNormal distribution of the masks is via independent sampling.
\subsubsection{Independent Sampling}
%From \cref{eq.logitN_one_dim} to obtain a logitNormal sample, it suffices to sample $z\sim \mathcal{N}(0, I_m) \in \mathcal{Z} = \mathbb{R}^m$ then apply the sigmoid function.
A common assumption is that the logitNormal samples originate from a factorized Normal distribution \cite{kocisky_semantic_2016}.
Thus, the learned parameters of the distribution are: the average $\mu \in \mathbb{R}^{m}$ and the diagonal coefficients of the covariance matrix $\sigma \in \mathbb{R}^{m} $, according to:
\begin{equation}
Y = \text{sigmoid}_{\lambda}(\mu + z \odot \sigma) \label{eq.iln} 
\end{equation}
where $\odot$ is the element-wise product and $ Y\in \mathbb{R}^{m}$. 
Note that, for feature selection on images, one aims at learning a binary mask and thus the latent space has the same dimension as the images, i.e. $m=n\times n$, then $z\in \mathbb{R}^{n\times n}$.

This sampling method has two main drawbacks.
First, the coordinates of $z$ are independent and so are the coordinates of $Y$, therefore such sampling scheme does not take correlations into account.
Also, the dimension of the sampling space $\mathcal{Z}$ is the same as $Y$ which might be prohibitive for large images.

We address both limitations in the following section, by considering the relations between the pixel values. 
%Indeed, in applications such as feature selection, we want to select the best subset of pixels for image reconstruction.
In that perspective, Group-Lasso selects variables among previously designed group of variables \cite{Yuan2006}, reflecting different aspects of the data.
Similarly, we want to select variables evidencing different facets of the signal to be observed. 
Indeed, finding the best subset of variables for the reconstruction implies to eliminate the redundancy in the signal and to explore the space of possible masks.
We propose to do so by selecting the variables in a correlated fashion, avoiding the selection of redundant information.

\subsubsection{Correlated Sampling:}
To palliate the limitations of independent sampling, we model the covariance between latent variables by learning linear combinations between the variables in the prior space $\mathcal{Z}$.
Besides, considering dependencies between latent variables, this mechanism reduces the dimension of the sampling space $\mathcal{Z}$, allowing for a better exploration of the latent space.
In order to generate correlated variables from a lower dimensional space, we investigate the following transformation: let $z \sim \mathcal{N}_d(0, I_d) \in \mathcal{Z}=\mathbb{R}^d$ with $d<<m$, $W \in \mathcal{M}_{m, d}(\mathbb{R})$ a weight matrix of size $m \times d$ and $b \in \mathbb{R}^{m}$ a real vector, then
\begin{equation}
Y=\text{sigmoid}_\lambda(Wz +b)\label{eq:binary_mask}
\end{equation} represents $m$-one dimension logitNormal laws due to the stability of independent Gaussian laws by addition.
However, the Normal law induced by $Wz+b$ has now a full covariance matrix and not only diagonal coefficient as in \cref{eq.iln}.
This reparameterization provides a simple way to sample correlated (quasi)-binary variables, even for high dimension latent space, i.e with $m$ large.

Compared to \cite{abid_concrete_2019}, our proposition offers a significant advantage for feature selection in images.
Indeed, let $G$ be the neural network aiming to reconstruct data $x$ from the selected variable.
With our proposition $G$ can access a sparse version of the original signal $Y \odot x$ and can thus leverage both the pixel values and their position in the image for reconstruction.
In \cite{abid_concrete_2019} only the selected feature values without structural information are available for the reconstruction.

\subsection{Parameterizing logitNormal Variables for Feature Selection} \label{subsec:param}
Now we have established how to compute correlated logitNormal variables following \cref{eq:binary_mask}, we detail our parameterization for learning.
Let $S: \mathcal{Z} \rightarrow [0,1]^{n\times n}$ be our sampling operator that generates a binary mask from a random sample $z$.
We consider two approaches to parameterize $S$ so that it follows a logitNormal law.
Our first proposition denoted vanilla parameterization directly optimizes $W$ and $b$ from \cref{eq:binary_mask}, while our second approach proposes to explore and optimize the spaces of linear combinations $W$ and biases $b$.
%To perform feature selection, we have to make our sampling operator $S$ not adapted to each data sample $x$, but rather adapted to the overall data distribution of $x\sim p_x$.

\begin{figure}[h]
  \centering
    \includegraphics[width=0.85\textwidth]{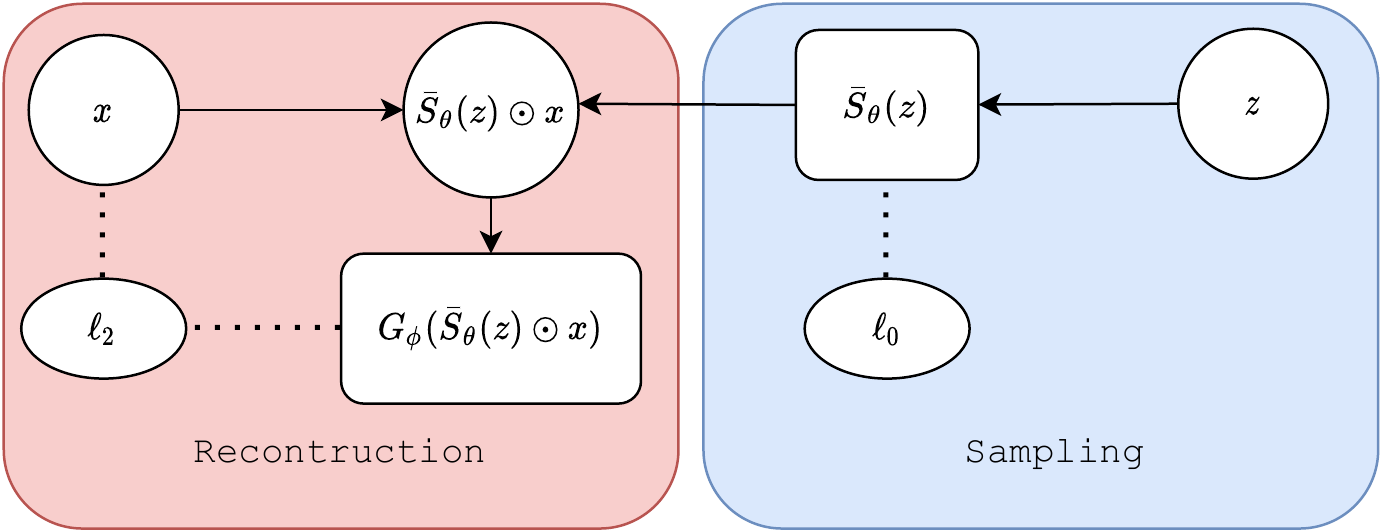}
    %\vspace{-0.2in}
    \caption{Algorithmic flow of our framework for feature selection for reconstruction.  $S_\theta(z)$ has a correlated logitNormal distribution. 
    We sample $z\sim \mathcal{N}(0,1)$. $\Bar{S}_\theta(z)$ defines the binary masks and $G_\phi$ estimate $x$ from ${x^{obs}=\Bar{S}_\theta(z) \odot x}$.}
    \label{fig:algo_flow}
    \vspace{-0.3cm}
\end{figure}

\subsubsection{Vanilla Parameterization:}
A simple approach is to parameterize $S$ as $S_\theta$ according to \cref{eq:binary_mask}.
Then, the optimized parameters are: ${\theta=(W, b)}$ with $W\in\mathcal{M}_{n\times n,d}(\mathbb{R})$ and $b\in \mathbb{R}^{n \times n}$.
This sampling process can be summarized by \cref{eq.vanilla}:
\begin{subequations}
\label{eq.vanilla}
\begin{empheq}[left={\empheqlbrace\,}]{align}
    &\text{Initialize } W\in\mathcal{M}_{n\times n,d}(\mathbb{R}),b \in \mathbb{R}^{n\times n} \nonumber \\
    &z \sim \mathcal{N}(0, I_d) \\
    &S_\theta(z) = \text{sigmoid}(W.z + b)
    \end{empheq}
\end{subequations}

In that case each variable in $S_\theta(z)$ follows a logitNormal law.
The selected variables are indicated for $S_\theta = 1$.
The optimization process allows two degrees of freedom ($b$ and $W$) for the control of the variance, of the covariance and of the average of the variables of the masks. 
Note that, this parameterization corresponds to a linear layer followed by a sigmoid activation so that besides tractability for the distribution of $Y$, it presents the advantage of a simple implementation.
Unlike \cite{abid_concrete_2019}, our proposition preserves the structure of the data.

\subsubsection{HyperNetworks Parameterization:} 
Aiming to learn a matrix $W$ and a bias vector $b$ that fully characterizes our logitNormal law as \cref{eq:binary_mask}, we leveraged in \cref{eq.vanilla} the stability of independent Gaussian law by addition.
However, the space of the linear combinations to be learned is high dimensional and structured, hence hard to learn.
Also, the optimization of the parameterization as \cref{eq.vanilla} is highly dependent on the initialization, as we optimize $W$ and $b$ from a (randomly) chosen start point.
Therefore, we want to be able to reach a wider space of parameters $W,b$.
To do so, we build on \cite{Karras2019} that successfully leverages latent code pre-processing with neural network in the context of adversarial learning for image generation, and \cite{Le2017} where a neural network generates the weights of another neural network to facilitate learning.
Therefore, instead of learning directly $W,b$ as in the vanilla approach we propose to learn to sample on the space of linear combination $W$ and biases $b$.
The core idea is to leverage neural networks expressivity to enrich the space of reachable matrices $W$ and vectors $b$ compared to the vanilla approach.
To do so we use the random sample $z$ to extract a representation vector $r\in\mathbb{R}^k$.
This representation $r$ serves as input to neural networks $F_b$, $F_W$ providing estimates of $W$ and $b$.
To sum up, in the HyperNetwork approach we learn a logitNormal law according to:
\begin{subequations}
\label{eq.hypernet}
\begin{empheq}[left={\empheqlbrace\,}]{align}
    &z  \sim \mathcal{N}(0, I_d), \ r = F_{rep}(z) \in\mathbb{R}^k ,\label{eq.rep} \\
    &W = F_W(r) \in \mathcal{M}_{n\times n, d}(\mathbb{R})  \label{eq.mat} , \\
    &b = F_b (r)\in \mathbb{R}^{n\times n} \label{eq.bias}, \text{ and finally:} \\
    &S_\theta(z) = \text{sigmoid}(F_b(r) + F_W(r).z) \label{eq.hyper_logit},
\end{empheq}
\medskip
\end{subequations}
Note that as desired $S_\theta(z)$ follows a logitNormal law  $\mathcal{L}\mathcal{N}(F_b(r), F_W(r)^T . F_W(r))$.
This proposition presents several advantages.
First, in \cref{eq.vanilla} $W$ is a randomly initialized weight matrix, then we only explore one trajectory of optimization from this (randomly chosen) starting point.
Also, instead of learning a distribution of masks, this parameterization learns a distribution of transport matrices and biases.
Therefore, both $F_W$ and $F_b$ stochastically explore a direction for each sample of $z$, providing more feedback with respect to the objective of feature selection for reconstruction.
This parameterization of $W$ and $b$ offers a way to explore efficiently the space of biases and linear combinations.
Also, because it rely on matrix multiplication, this procedure is computationally barely less efficient than the naive one when $F_W$ and $F_b$ are small neural networks.

We show experimentally the superiority of this approach in \cref{sec.exp}.

\subsection{Sparsity Constraint: $\ell_0$-Relaxation} \label{sec.mask_sparsity}
We detail our approach promoting sparsity.
Frequently, sparsity in regression settings is enforced thanks to a $\ell_1$ penalty on the parameters.
However, $\ell_1$ approaches may suffer from a shrinking effect due to ill-posedness as detailed in \cref{sup.l1}.
%However in our setting of feature selection for reconstruction, a $\ell_1$ penalty on the masks makes the problem ill posed, see \cref{sup.l1}.
%Moreover, we want to avoid the shrinking effect induced by $\ell_1$.
Consequently, we introduce an alternative approximation of the $\ell_0$-formulation better suited to our feature selection application: we minimize the expected $\ell_0$-norm, i.e the probability of each variable in our binary mask to be greater than $0$.
Thus, we need a non zero probability of sampling $0$ which is not the case with the current scheme.
Accordingly, we introduce a stretching scheme to obtain a non-zero mass at points ${0}$ and ${1}$ while maintaining differentiability.
% IDEE comparer les gradient de la l0 et de la l1 ?

\subsubsection{Stretched Distribution} To create a mass at ${0}$, we proceed as in \cite{louizos_learning_2017}.
Let $Y \in [0,1]^{m}$ be a logitNormal variable, $\gamma<0$ and $\eta >1$ and $HT$ be the hard-threshold function defined by $HT(Y) = \min(\max(Y, 0),1)$, the stretching is defined as:
\begin{equation}
\Bar{Y} = HT\{(\eta - \gamma) Y + \gamma\} \label{eq.stretching}
\end{equation}
Thanks to this stretching of our distribution, we have a non zero probability to be zero, i.e ${\mathbb{P}(\Bar{Y}=0)>0}$ and also ${\mathbb{P}(\Bar{Y}=1)>0}$.
Further details are available in supplementary \cref{sec.stretching}.
We can now derive a relaxed version of the $\ell_0$-norm penalizing the probability of the coordinates of $\Bar{Y}$ to be greater than $0$.

\subsubsection{Sparsity Constraint:}
Let $L_0(\Bar{Y})$ the expected $\ell_0$-norm of our stretched output $\Bar{Y}$.
Using the notation in \cref{eq.stretching}, we have:
\begin{equation}
L_0(\Bar{Y}) = \mathbb{E}[\ell_0(\Bar{Y})] = \sum_{i=1}^{m} \mathbb{P}(\Bar{Y}_i> 0)
= \sum_{i=1}^{m} 1 - F_{Y}\big(\frac{-\gamma}{\eta - \gamma}\big), \label{l0.eq}
\end{equation}
where $F_{Y}$ denotes the cumulative distribution function (CDF) of $Y$.
This loss constrains the random variable $Y$ to provide sparse outputs as long as we can estimate $F_{Y}$ in a differentiable way.
In the case of the logitNormal law, we maintain tractability as $Y$ satisfies \cref{eq:binary_mask} or \cref{eq.iln}.
Thus, for our $m$-dimensional logitNormal law defined as in \cref{eq:binary_mask}, we have:
\begin{equation}
 L_0(\Bar{Y}) = \sum_i 1 - \Phi\Big(\frac{\log (\frac{-\gamma}{\eta}) - b}{\sqrt{\sum_j W_{j,i}^2}}\Big)
\label{eq.l_0_ln}, 
\end{equation}
where $\Phi$ is the CDF of the unitary Normal law.
Detailed computations are available in supplementary materials \cref{sec.l0_ln}.
Minimizing \cref{eq.l_0_ln} promotes sparsity in the law of $Y$ by minimizing the expected true $\ell_0$-norm of the realisation of the random variable $Y$.
We have developed a constraint that promotes sparsity in a differentiable way. Now we focus on how to learn efficiently the parameters of our correlated logitNormal law.

\subsection{Reconstruction for Feature Selection} \label{subsec:reconstruction}
We have designed a sparsity cost function and detailed our parameterization to learn our sampling operator, we focus on the downstream task.
Consider data $(x_i, y_i)_{i\in[1..N]}$, consisting in paired input $x$ and output $y$. Feature selection consists in selecting variables in $x$ with a mask $s$, so that the considered variables: $s\odot x$ explain at best $y$.
Let $G$ be a prediction function and $\mathcal{L}$ a generic cost functional, feature selection writes as:
\begin{equation} \label{eq.loss_fs}
    \min_{s,f} \mathbb{E}_{x,y} \ \mathcal{L}\big(G(s\odot x), y\big)
    \ \ \text{s.t } ||s||_0 < \lambda,
\end{equation}

In this work we focus on a reconstruction as final task, i.e $y=x$.
Besides the immediate application of such formulation to optimal sensors placement and data compression, reconstruction as downstream task requires no other source of data to perform feature selection.
Naturally, this framework is adaptable to classification tasks.
As a sparse auto-encoding technique, feature selection with a reconstruction objective aims at minimizing the reconstruction error while controlling the sparsity.
In this case $G_\phi:\mathbb{R}^{n\times n} \rightarrow \mathbb{R}^{n\times n}$ is our reconstruction network (of parameter $\phi$) taking as inputs the sparse image.
The feature selection task with an $\ell_2$-auto-encoding objective writes as:
\begin{equation} \label{eq.loss_l2}
    \min_{\theta, \phi} \mathbb{E}_x||G_\phi(\Bar{S}_\theta(z) \odot x) - x||_2 + \lambda_{sparse} L_0(\Bar{S}_\theta(z))
\end{equation}
A schematic view of our proposition, illustrating the sampling and the reconstruction component is available in \Cref{fig:algo_flow}.
An algorithmic description in the vanilla case (\cref{eq.vanilla}) is available in supplementary \cref{sup.algorithm}.

%However, $\ell_2$ autoencoding is known for bluriness in the output \cite{isola_image--image_2016}. 
%Indeed, consider $s$ fixed.
%The optimum of \cref{eq.loss_l2} is ${G_\phi(\Bar{S}_\theta(z) \odot x) = \mathbb{E}(x|\Bar{S}_\theta(z) \odot x)}$.
%If the conditioning $\Bar{S}_\theta(z) \odot x$ is very sparse or weakly informative, $G_\phi$ will render blur results, averaging among possible outcomes.

\begin{comment}
\paragraph{Via cGan}
To palliate this drawback, we consider a conditional GAN (cGAN) approach like in \cite{isola_image--image_2016} solving the issues raised by $\ell_2$-auto-encoding.
\begin{equation}
    \min_{s, \phi} \max_{\psi} \lambda_{\ell_1}||x - G_\phi(\Bar{S}_\theta(z) \odot x)||_1 + \mathcal{L}_{gan}(G_\phi, D_\psi) + \lambda_{sparse} L_0(\Bar{S}_\theta(z)) ,
\end{equation}
with: 
\begin{equation}
    \mathcal{L}_{gan} = \log D_\psi\big(x,x \odot \Bar{S}_\theta(z)\big)
+ \ \mathbb{E}_{z,x} \log \big\{ 1 - D_\psi\big( G_\phi(x \odot \Bar{S}_\theta(z)), x \odot \Bar{S}_\theta(z)\big) \big\}
\end{equation}

Note that our binary mask distribution is optimized for image reconstruction through an end to end training. 
\end{comment}

\section{Experiments} \label{sec.exp}
We provide experimental results on 3 datasets: MNIST, CelebA and a geophysical dataset resulting from complex climate simulations \cite{ipsl_simul,sepulchre2019}.
We use the traditional train-test split for MNIST and a 80-20 train-test split for the other datasets.
The geophysical dataset is composed of surface temperatures anomalies (deviations between average temperature at each pixel for a reference period and observations) and contains 21000 samples (17000 for train).
The data have both high (Gulf stream, circum-polar current ...) and low frequencies (higher temperature in the equatorial zone, difference between northern and southern hemispheres ...) that need to be treated accurately due to their influence on the Earth climate.
Accuracy in the values of reconstructed pixel is then essential for the physical interpretation.
These dense images represent complex dynamics and allow us to explore our method on data with crucial applications and characteristics very different from the digits and faces.

\subsection{Experimental and Implementation Details}
%We detail our experimental setting and implementation choices.

\subsubsection{Baselines} Besides our models Vanilla logitNormal, denoted \textit{VLN}, and its hyper-networks couterpart denoted \textit{HNet-LN}, we consider as competing methods the following approaches:
\begin{enumerate}
    \item Concrete-Autoencoder \cite{abid_concrete_2019} denoted \textit{CAE}.
    \item To assess the relevance of our correlated proposition, we investigate a binary mask approach based on the independent logitNormal mask that corresponds to equation \cref{eq.iln} denoted \textit{ILN},
    \item Another independent binary mask method based on the concrete law \cite{maddison_concrete_2016}, see supplementary materials \cref{sup.concrete}, denoted \textit{SCT}.
\end{enumerate}

\subsubsection{Implementation Details}
For all binary mask based methods, we use a Resnet for $G_\phi$, \cite{he_deep_2015} following the implementation of \cite{isola_image--image_2016}.
$F_{rep}$, $F_W$ and $F_b$ are two layers MLP with leaky relu activation.
For CAE, because the structure of the data is lost in the encoding process, we train $G_\phi$ as a MLP for MNIST and a DcGAN for geophysical data and CelebA.
Thorough experimental details are available in \cref{sup.details}. The code is available at: \url{https://github.com/JeremDona/feature_selection_public}

\subsubsection{Removing Randomness:}
All masked based algorithms learn distributions of masks.
To evaluate the feature selection capabilities, we evaluate the different algorithms using fixed masks.
We rely on \cref{prop1} to remove the randomness during test time.
Let $S^0_\theta$ be the 0-temperature distribution of the estimated $S_\theta$.
We first estimate the expected $\ell_0$-norm of the 0-temperature distribution: $L_0(S^0_\theta)$.
We then estimate two masks selecting respectively the $10 \times \lfloor \frac {L_0(S^0_\theta)}{10} \rfloor$ and $10 \times \lceil \frac{L_0(S^0_\theta)}{10} \rceil$ most likely features (rounding $L_0(S^0_\theta)$ up and down to the nearest ten).
This method has the advantage of implicitly fixing a threshold in the learned mask distribution to select or reject features.
More details on the method are available in \cref{sup.practice}.

We now illustrate the advantage of selecting features in a correlated fashion.

\subsection{Independent vs Correlated Sampling Scheme:}
\subsubsection{Is a Covariance Matrix Learned ?}
\begin{figure}
    \vspace{-1em}
    \centering
    \includegraphics[width=0.7\textwidth]{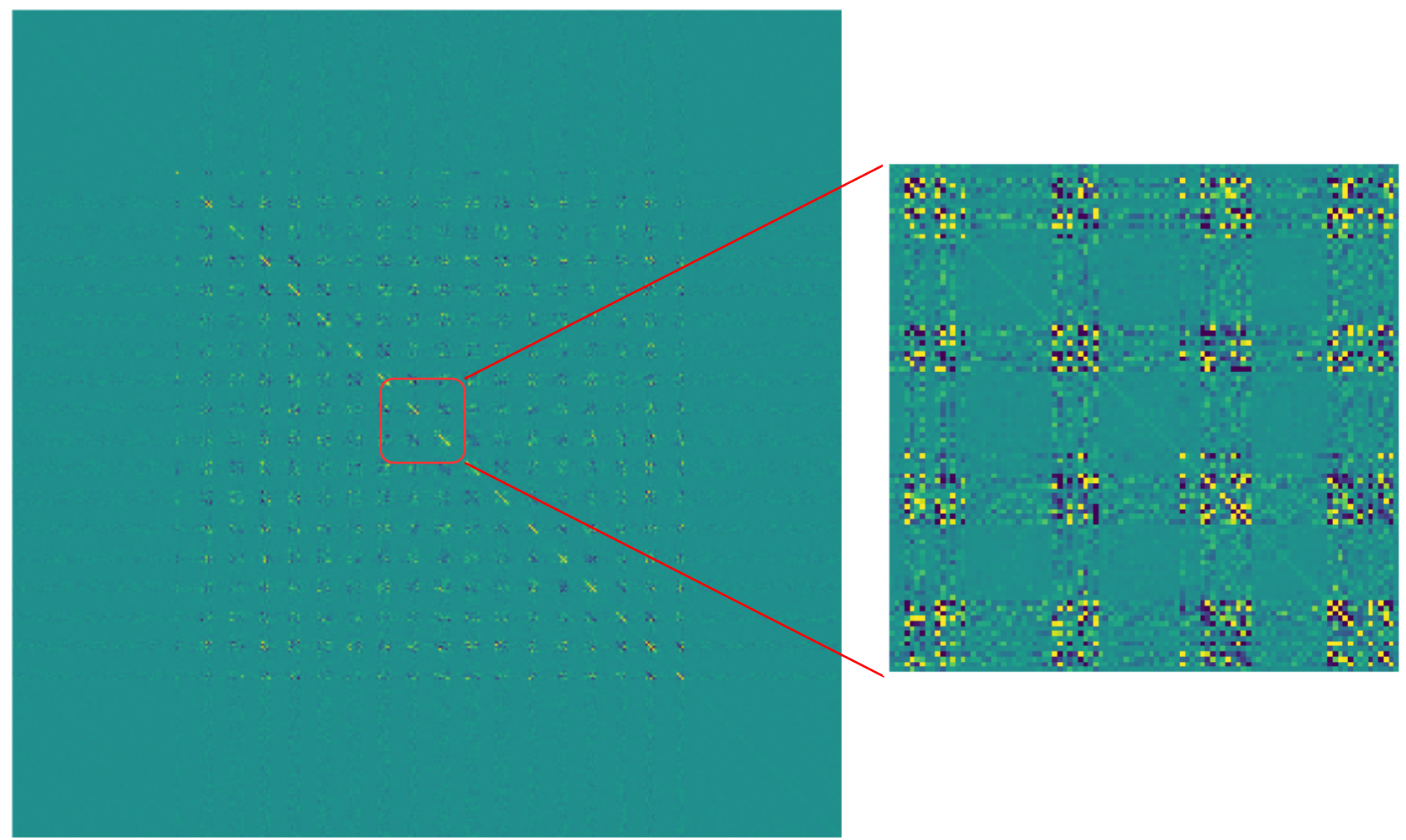}
    \caption{Covariance matrix learned with \cref{eq.vanilla}, with $\approx 30$ pixels selected. Yellow values indicates high positive covariance, blue ones low negative covariance}
    \label{fig:covar}
\end{figure}

Because we model the local dependencies in the sampling by learning linear mixing of latent variables $z$, we first verify the structure of the covariance matrix.
\Cref{fig:covar} reports the learned covariance matrix of the sampling for MNIST dataset using \cref{eq.vanilla} method.
Besides the diagonal, extra-diagonal structures emerge, revealing that local correlations are taken into account to learn the sampling distribution.

\subsubsection{Independent Sampling Does not Choose}
\begin{figure}[h]
\centering    \includegraphics[width=0.75\textwidth, trim=4.2cm 6cm 4cm 3.7cm,clip]{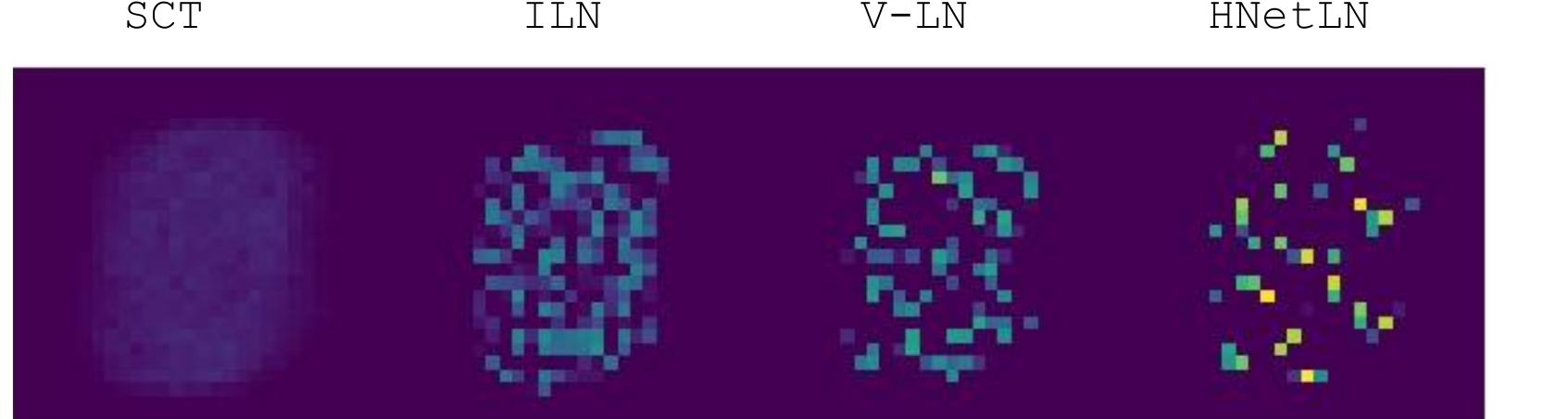}
\caption{Masks empirical distribution for competing binary masks algorithms on the MNIST datasets for about 30 features in the sampled mask}
\label{fig:mnist_mask_distribution}
\end{figure}

We show in \cref{fig:mnist_mask_distribution} the empirical average of the sampled masks for each masked base competing algorithm where all algorithms were trained so that at $L_0(S^0_\theta) \approx 30$.
\Cref{fig:mnist_mask_distribution} clearly shows that concrete base algorithm (SCT) and in a lesser sense (ILN) do not select features, but rather put a uniformly low probability to sample pixels in the center of the image.
This means that both algorithms struggle at discriminating important features from less relevant ones.
On the other hand, our correlated propositions, Vanilla logitNormal (V-LN, \cref{eq.vanilla}) and particularly the hyper-network approach (HNetL, \cref{eq.hypernet}) manage to sparsify the distribution prioritizing the selection of important pixels for reconstruction.

\subsection{Feature Selection and Reconstruction}

We now quantitatively estimate the impact of our choices on the reconstruction error on the various datasets. First, the mean squared error reconstruction results from \cref{tab.mse} tells us that considering the spatial structure of the data enhances reconstruction performance. 
Indeed, mask based methods consistently over-perform CAE where the data structure is linearized in the encoding process.
Furthermore for mask based method, correlated sampling (row V-LN and HNet-LN) also consistently improves over independent sampling based method (row ILN and SCT).
Finally, our hyper-network HNet-Ln proposition also improves over the vanilla approach validating our proposition. 
Samples for all datasets are available in supplementary \cref{sup.samples}
%This reduction in the reconstruction comes from the lower variance (see \cref{fig:mnist_mask_distribution}) of correlated schemes, making $G_\phi$ adapted to much lower samples of.
\begin{table}
\caption{Average Reconstruction Error (MSE) on MNIST, Climate and CelebA datasets for all considered baselines}\label{tab.mse}
\centering
\begin{tabular}{p{0.25\textwidth}P{0.07\textwidth}P{0.07\textwidth}P{0.07\textwidth}P{0.07\textwidth}P{0.07\textwidth}P{0.07\textwidth}P{0.07\textwidth}P{0.07\textwidth}P{0.07\textwidth}}
\toprule
& \multicolumn{3}{c}{MNIST}& \multicolumn{3}{c}{Climat} & \multicolumn{3}{c}{CelebA} \\ \midrule
\# Features   & 20  & 30 & 50 & 100 & 200 & 300 & 100 & 200 & 300 \\ \midrule
CAE  & 3.60 & 3.05 & 2.40  & 
2.07 & 1.98  & 1.96 
& 7.65 & 6.42 & 5.7 \\ \midrule
ILN & 3.67 & 2.41  & 1.41 
&  1.44 & 1.05 & 0.83
& 7.1 & 2.56 & 1.87    \\
SCT & 3.72 & 3.61 & 2.60
& 2.20 & 1.89 & 1.51 
& 7.99  & 3.31  & 2.44 \\
VLN (Ours)  & 3.22 & 2.19 & 1.33 
& \bfseries 1.11 &  \bfseries 0.93 & 0.79 
& 3.11 & 1.96 &  1.50 \\
HNet-Ln (Ours) & \bfseries 2.15 & \bfseries 1.53 & \bfseries 1.06 
& 1.78 &  0.96 & \bfseries 0.60 
&\bfseries  2.81 &\bfseries  1.7 & \bfseries 1.46 \\ \bottomrule
\end{tabular}
\end{table}

\subsection{Quality of the Selected Features: MNIST Classification}
We now assess the relevance of the selected features of our learned masks on another task.
To do so, for each learned distribution we train a convolutional neural network, with a DcGAN architecture on MNIST classification task.
Here also, the randomness in test set is removed.
For each mask we run $5$ experiments to account for the variability in the training.
Classification results reported in \cref{tab.mnist_classif} indicate that both our correlated logitNormal propositions consistently beat all considered baselines, validating our choices to learn a sampling scheme in a correlated fashion.
Indeed, our propositions systematically reach the lowest minimum and average classification error.

\begin{table}[h]
\caption{Classification error in percent for MNIST on test set for all considered baselines. Minimum and average are taken over 5 runs.} \label{tab.mnist_classif}
\centering
\begin{tabular}{p{0.23\textwidth}P{0.1\textwidth}P{0.1\textwidth}P{0.1\textwidth}P{0.1\textwidth}P{0.1\textwidth}P{0.1\textwidth}}
\toprule
\# Features    & \multicolumn{2}{c}{20} & \multicolumn{2}{c}{30} & \multicolumn{2}{c}{50} \\ \midrule
Metric & Min & Mean & Min & Mean & Min & Mean \\ \midrule
CAE & 24.4 & 31.64 & 
8.89 & 19.60  & 
5.45 & 6.65\\
ILN & 21.58 & 28.26 
& 7.96 & 16.63 & 
4.17 & 5.33 \\
SCT & 20.88 & 32.79 & 
9.49 & 18.22
& 4.11 & 6.77 \\
VLN (Ours) & \bfseries 12.15 & \bfseries 24.74 & 
\bfseries 6.38 & \bfseries 15.07 
& 3.32 & \bfseries 4.67 \\
HNet-LN (Ours) & 19.23 & 25.07 
& 7.24 & 17.80  & 
\bfseries 2.84 & 6.45  \\ \bottomrule
\end{tabular}
\end{table}

\subsection{Extension: cGAN}
\begin{figure}[h]
    \centering
    \includegraphics[width=1\textwidth, trim=0cm 8cm 1.7cm 2.12cm,clip]{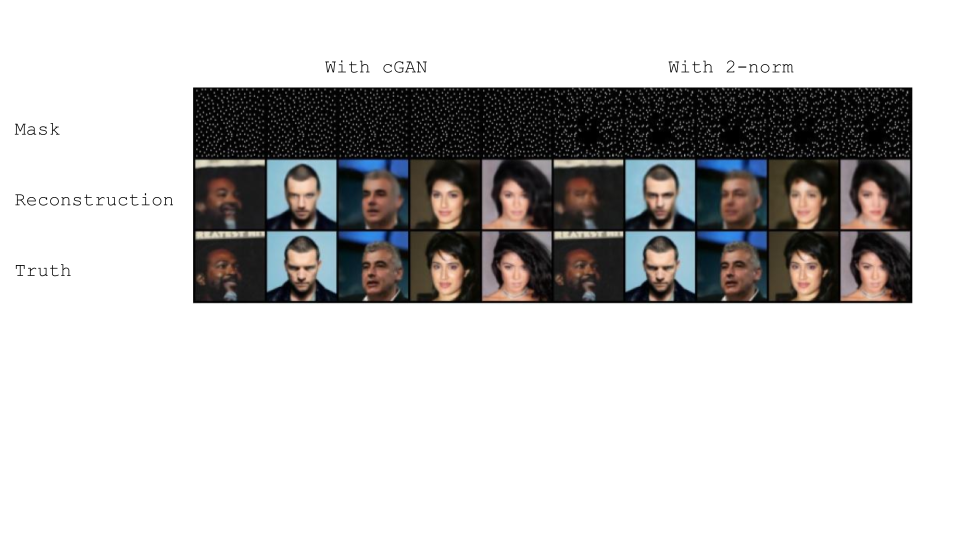}
    \caption{Examples of masks (first row), reconstructions (second row) and true data (last row) for CelebA dataset using either a cGAN (4 first columns) or simple auto-encoding (4 last columns) for 200 selected features. Best viewed in color.}
    \label{fig:cgan}
\end{figure}
We detailed in the previous experiments feature selection results obtained thanks to an $\ell_2$-auto-encoding approach.
This choice was motivated because in physical measurement all points are equals: we don't want to favor the reconstruction of some part of the image while neglecting another.
However, for images such as CelebA all points are not equal: the face part of the image being more interesting than the background.
Indeed, a realistic reconstruction can be preferred to a well reconstructed background.
Moreover, $\ell_2$-auto-encoding suffers from blur in the reconstruction.
In that perspective, we can leverage conditional generative adversarial networks (cGAN) approaches \cite{mirza_conditional_2014,isola_image--image_2016} that solves the blurriness occurring in $\ell_2$-decoding.
We implement the cGAN approach of \cite{isola_image--image_2016}.
\Cref{fig:cgan} illustrates that despite both method show good reconstruction, the cGAN approach on CelebA enables a stronger focus on faces facilitating realistic reconstruction.
We refer to \cref{sec.cgan} for more details and samples.

\section{Conclusion}
In this work, we formulate the feature selection task as the learning of a binary mask. 
Aiming to select features in images for reconstruction, we developed a novel way to sample and learn a correlated discrete variable thanks to a reparameterization of the logitNormal distribution.
The proposed learning framework also preserves the spatial structure of the data, enhancing reconstruction performance.
We experimentally show that our proposition to explore the space of covariance matrices and average vectors as in \cref{eq.hypernet} is efficient providing us with a sampling with lower variance.
Finally, we experimentally evidenced the advantage of learning a correlated sampling scheme instead of independent ones.

\bibliographystyle{splncs04}

\bibliography{references.bib}
\begin{comment}

\end{comment}

\newpage
\section{Supplementary Material}
\subsection{The logitNormal Distribution:}
\label{sup.ln_muli}
The logitNormal distribution defines a probability distribution over the simplex, see \cite{aitchison_logistic-normal_1980}
Initially introduced to describe compositional data \cite{aitchison_statistical_1982}, it is defined has:
\begin{definition}{LogitNotmal}
\label{def.logitn}
Let X be random variable defined over $\mathbb{R}^n$ such that $X \sim \mathcal{N}(\mu, \sigma)$. Then, consider the following transformation:
$$ Y_{-n} = e^X / (1+\sum_{j=1}^n e^{X_j}) \text{, and } Y_{n+1} = 1 - \sum_{j=1}^n Y_j$$
Then the vector $Y=(Y_1,...Y_{n+1})$ follows a logitNormal distribution denoted $\mathcal{L}\mathcal{N}(\mu, \sigma)$ and is defined over the $\mathbb{R}^{n+1}$ simplex.
Moreover, $Y$ admits a density and can be found in \cite{aitchison_logistic-normal_1980}.
\end{definition}

If $Y \sim \mathcal{L}\mathcal{N}(\mu, \sigma)$, it defines a probability distribution over the simplex which makes it practical to model compositional data, i.e \say{data where the involved data forms some sort of proportion of a whole} \cite{aitchison_statistical_1982}.

\subsection{Reparametrizing the logitNormal Distribution:}
Using \Cref{def.logitn} of the logitNormal distribution, we can use the reparameterization trick in order to learn the parameters of a logitNormal law from samples of Normal law.

\begin{theorem}{Reparameterization:}
Let $X=(X_i)_{i\leq n}$ such that $X_i \sim \mathcal{N}(0, 1)$ and all $X_i$ are iid (X $\in \mathbb{R}^n$), $W \in \mathcal{M}_{m\times n}(\mathbb{R})$, and $b \in \mathbb{R}^m$, then :
\begin{align}
& Y_{-n}= \exp(WX + b) / (1+ \sum_i \exp({W_i.X + b_i})) \nonumber\\
& Y = (Y_{-n}, 1 - \sum_{j=1}^n Y_j) \\
&Y \sim \mathcal{L}\mathcal{N}(b, \Sigma)
\end{align}
\end{theorem}
This comes from the simple fact that an affine transformation of i.i.d. $\mathcal{N}(0,1)$ follows also a Normal law, which co-variance matrix can be expressed through the matrix of linear weights. Moreover, this advantageously correspond to a neural network layer with an extended sigmoidal function.

\subsection{Proof For 0-Temperature} \label{proof.prop1}
Here we prove the convergence of the reparameterization of the logitNormal law for the zero temperature.
%\vspace{-1cm}
\begin{proof} 
Let $(\lambda_n)_{n\geq0}$ be a positive sequence decreasing towards $0$.
We prove the 0-temperature convergence for $z \sim \mathcal{N}(\mu, \sigma)$.
Let $Y_n=\text{sigmoid}_{\lambda_n}(z)$. We investigate the convergence in distribution of $Y_n$ towards a Bernoulli distribution.
Let $f$ be a continuous bounded function.
We have:
\begin{align*}
    \mathbb{E}(f(Y_n)) &= \int_0^1 f(Y_n)dP_{Y_n} = \int_{\mathbb{R}} f(\text{sigmoid}_{\lambda_n}(z)) dP_z \\
    &= \int_{\mathbb{R}}f(\text{sigmoid}_{\lambda_n}(z)) \frac{1}{\sqrt{2 \pi}\sigma}\exp^{-\frac{1}{2} (\frac{z-\mu}{\sigma})^2} dz
\end{align*}
We first have point-wise convergence of the sequence of function inside the integral.
Indeed,

If $z>0, \lim_{n\rightarrow\infty} \text{sigmoid}_{\lambda_n}(z) = 1$.

If $z<0, \lim_{n\rightarrow\infty}\text{sigmoid}_{\lambda_n}(z) = 0$. We have:
\begin{equation*}
\lim_{n\rightarrow\infty} f(\text{sigmoid}_{\lambda_n}(z)) \frac{1}{\sqrt{2 \pi}\sigma}\exp^{-\frac{1}{2} (\frac{z-\mu}{\sigma})^2} = \frac{1}{\sqrt{2 \pi}\sigma}f(\delta_{z>0})\exp^{-\frac{1}{2} (\frac{z-\mu}{\sigma})^2} 
\end{equation*}
The domination is verified using the function:
\begin{equation*}
g(z)=\frac{1}{\sqrt{2 \pi}\sigma}||f||_\infty \times \exp^{-\frac{1}{2} (\frac{z-\mu}{\sigma})^2}
\end{equation*}
We can finally apply the theorem of dominated convergence:
\begin{align*}
\lim_{n\rightarrow\infty}\mathbb{E}(f(Y_n)) &=
\mathbb{E}(\lim_{n\rightarrow\infty}f(Y_n)) \\
&= \int \frac{1}{\sqrt{2 \pi}\sigma}f(\delta_{z>0})\exp^{-\frac{1}{2} (\frac{z-\mu}{\sigma})^2} dz \\
& = \frac{1}{\sqrt{2 \pi}\sigma}f(0) \int_{-\infty}^0 \exp^{-\frac{1}{2} (\frac{z-\mu}{\sigma})^2} dz + \frac{1}{\sqrt{2 \pi}\sigma}f(1)\int_{0}^{+\infty} \exp^{-\frac{1}{2} (\frac{z-\mu}{\sigma})^2} dz \\
& = f(0) \Phi(-\frac{\mu}{\sigma}) + f(1)(1- \Phi(-\frac{\mu}{\sigma})) \\
& = \mathbb{E}_{b\sim{\mathcal{B}(1-\Phi(\frac{-\mu}{\sigma}))}}f(b),
\end{align*}
where $\mathcal{B}$ denotes Bernoulli distribution.
Finally, we can conclude that $Y_n$ converges in law towards a Bernoulli distribution such that: $Y_n\rightarrow\mathcal{B}\big(1-\Phi(\frac{-\mu}{\sigma})\big)$

\end{proof}

\subsection{Proof $L_0$-logitNormal:} \label{sec.l0_ln}
\begin{align*}
 L_0(S_\theta(z)) &= \sum_i 1 - \mathbb{P}(\Bar{z} \leq 0)\\
   &= \sum_i 1 - \mathbb{P}(\text{sigmoid} (Wz+b)\leq -\gamma / (\eta- \gamma)) \\
   &= \sum_i 1 - \mathbb{P}(W_i.z \leq \log (\frac{-\gamma}{\eta}) - b)\\
   & \text{as $W_i.z$ has a normal law $\mathcal{N}(0, \sqrt{\sum_j w_{j,i}^2})$} \\
   &= \sum_i 1 - \Phi(\frac{\log (\frac{-\gamma}{\eta}) - b}{\sqrt{\sum_j W_{j,i}^2}})
\end{align*}

\subsection{Ill poseness of the $\ell_1$-formulation:} \label{sup.l1}
Consider the auto encoding setting with a $\ell_1$-norm instead of the derived $L_0$. The optimization problem is:
\begin{equation}
    \mathcal{L}_{\ell_2}=\lambda_{\ell_2}\mathbb{E} _{x\sim p_x}||x - G_\phi(S_\theta(z) \odot x)||_2 + \lambda_{s}.L_1(S_\theta)\label{eq.l1}
\end{equation}

Let $(G^*_\phi, S^*_\theta)$ be an optimal solution, i.e that realizes the minimum of the above optimization cost function.
Then, consider: ${S_2 = S^*_\theta / 2}$ and $G_2$ defined as $G_2(x) = G^*_\theta(2*x)$.
Then the MSE term of \cref{eq.l1} for the couple $(G_2, S_2)$ is equivalent as the one with $(G*_\phi, S*_\theta)$, however the $\ell_1$-norm of ($S_2$) is lower. Therefore $(G^*_\phi, S^*_\theta)$ is not optimal and the problem of \cref{eq.l1} is ill-posed.
However, note that, in the case of binary vectors, $\ell_0$-norm and $\ell_1$-norm are equals.

\subsection{On the Stretching Scheme:} \label{sec.stretching}
We initially start from a distribution $p$ that lives in $[0, 1]$ and need to transform it in order to obtain a non zero probability of sampling $0$ while maintaining both tractability and differentiability.
We denote this function $f$. 
We need $f^{-1}(0)$ to be a non-zero measure set of the original support.
In other words, we need $f$ to be a surjection, and $f^{-1}(0)$ to be Lebesgue measurable with a non zero mass. 
Instead of the $HT$ function we could have used a stretched $relu$ function.
One significant advantage of the chosen function is that it also creates a non-zero probability of sampling $1$ therefore enforcing the binary behaviour of our masks.
Unbalanced binary scheme can also be investigated in future works.
Indeed one can think of creating a higher portion of the stretched distribution above one, enforcing the binary behaviour of the mask.

\subsection{Algorithm} \label{sup.algorithm}
We present here the algorithm for the proposed logitNormal based feature selection algorithm.
\begin{algorithm} 
\SetAlgoLined
\KwResult{Converged $S$ $G_\phi$}
Initialize $\theta = (W, b)$ and $G_\phi$ \\
 \While{Convergence not reached}{
 sample batch $x=(x_1,...x_n)$\;
 and $z=(z_1,...,z_n)$, such that $z_i\sim \mathcal{N}(0, I_d)$\;
 
Compute $S_\theta(z)= \text{sigmoid}_\lambda(Wz+b)$\; and the observations $x^{obs} = \Bar{S}_\theta(z) \odot x$\;

Estimate reconstruction $\hat{x}=G_\phi(x^{obs})$\;
 
$L = ||x-\hat{x}||_2 + \lambda_{sparse} L_0(\Bar{S}_\theta(z))$\;

Update $\phi$ and $\theta$:
\begin{align*}
    \phi &\leftarrow \phi - \frac{\partial L}{\partial \phi} \\
    \theta &\leftarrow \theta - \frac{\partial L}{\partial \theta}
\end{align*}
 }
 \caption{Differentiable Feature Selection}
  \label{algo.general}
\end{algorithm}

\subsection{Practical Consideration on the Temperature:}
As duely noted by \cite{maddison_concrete_2016}, the temperature in sigmoid activation plays a crucial role in the training.
This remark holds for our work.
Indeed, in our work decreasing the temperature in the sigmoid, amounts to increase the variance and the absolute value of the average of the initial Gaussian distribution.

Also aiming at approximating binary distribution, we don't want any interior mode as in the green curve depicted in \cref{fig:std}: $\mathcal{L}\mathcal{N}(0, 1)$ has an 
interior maximum point.
This case is not acceptable for the approximation of Bernoulli random variable as, it could allow a leakage of information, i.e the distribution is not approximating a binary distribution anymore.
Therefore, during training one should ensure that the learned distribution has no interior maxima.
Fortunately, it suffices to sufficiently decrease the temperature $\lambda$ of the sigmoid in order to recover two modes at $0$ and $1$.
Indeed, decreasing sufficiently the temperature in the sigmoid pushes the interior maximum towards the edges.
In practice, we observe that initializing our $W$ so that $W.z$ with a variance higher than $0.5$ with a temperature of $\lambda=0.3$ suffices.

\subsection{Removing the Randomness} \label{sup.practice}
Both our propositions of \cref{eq.vanilla} or \cref{eq.hypernet} estimates distribution in the spaces of binary variables.
To collapse the distribution, one can take advantage of \cref{prop1} and select the $K$ desired number of features.
One can also, empirically select the $K$ features the mask with the highest probability to be selected.
Both approaches lead to similar results in practice.
Note that in both cases, if $K$ is far from the observed number of pixel, the selected features may not be the best subset of the learned distribution.

In practice, we chose to collapse the distribution using \Cref{prop1}:
We first estimate the expected $\ell_0$-norm of the distribution, which equals to $\sum (1-\phi(-\frac{\mu_i}{\sigma_i})$).
Let $L_0$ be the value of the expected $\ell_0$-norm of our learned distribution.
We then select two masks made of the most likely features to be selected: the first one has $L_0$ rounded \textit{down} to the nearest ten pixels.
The other one has $L_0$ rounded \textit{up} to the nearest ten selected pixels.
Note that, for SCT baseline, we use a property similar to \Cref{prop1} for the concrete distribution, available in \cite{maddison_concrete_2016}.

\subsection{Concrete Law} \label{sup.concrete}
Introduced by \cite{maddison_concrete_2016} to approximate discrete variables, binary concrete random variable is defined as follows:
\begin{align*}
u &\sim \mathcal{U}([0,1]) \nonumber \\
G &= \log(u) -\log(1-u)\nonumber \\
X &= \text{sigmoid}\Big(\frac{\log(\alpha) + G}{ \lambda}\Big) \label{gumbel.eq},
\end{align*}
And X follows a relaxed binary concrete law.

\subsection{Experimental Details} \label{sup.details}
All experiments were trained on Titan XP GPU via using Pytorch framework and mixed precision training.
For all experiments the expected $\ell_0$-norm is normalized by the number of pixels in the signal.
Also for all algorithms trained using correlated logitNormal approach, the dimension of $z$ is $16$, i.e. $ z\sim \mathcal{N}(0, I_{16})$.

For all mask based methods, $G_\phi$ is a resent following the implementation of \cite{isola_image--image_2016} with 2 residual blocks and $16$ filters.

\subsubsection{Mnist}
All masked based algorithms were trained using ADAM optimizer with $\beta =(0.9, 0.99)$ and a learning rate of $2.10^{-4}$ for $550$ epochs with batch size 256.
CAE method was trained for $1400$ epochs with a temperature decreasing form $10$ to $0.01$ following recommendation of the authors.

\subsubsection{Climate Data}
All masked based algorithms were trained using ADAM optimizer with $\beta =(0.9, 0.99)$ and a learning rate of $2.10^{-4}$ for $550$ epochs with batch size 128.
CAE method was trained for $1400$ epochs with a temperature decreasing form $10$ to $0.01$ following recommendation of the authors.

\subsubsection{CelebA}
All masked based algorithms were trained using ADAM optimizer with $\beta =(0.9, 0.99)$ and a learning rate of $2.10^{-4}$ for $140$ epochs with batch size 128.
CAE method was trained for $400$ epochs with a temperature decreasing form $10$ to $0.01$ following recommendation of the authors.

\subsubsection{Hyperparameters Search}
Except for CAE where the number of selected features is a structural constraint, we search the hyperparameter space by sampling from the interval $[10^{-2}; 1]$ discretized by steps of $3.10^{-2}$.
For all dataset, the CAE method was trained with a decreasing temperature from $10$ to $0.01$ following the guidelines of the authors \cite{abid_concrete_2019}.
For the mask method based on the concrete distribution the temperature of the sigmoid was set to $\lambda=2/3$ following the recommendation of \cite{maddison_concrete_2016}.
For logitNormal based algorithm, the temperature was fixed to $\lambda=0.3$.

\subsubsection{Initialization}
For all mask based methods, we chose the initialization parameters so that the resulting distribution of the each variable in the mask is symmetrical, with as many chances to be sampled than to be rejected, i.e. for all variable $i$ in the masks: $\mathbb{P}(S_\theta(z)_i<\epsilon)\approx \mathbb{P}(S_\theta(z)_i>1-\epsilon)\approx 0.2 $.
That way, all distribution can explore the space of binary masks.
Also, in order to verify whether a covariance matrix is learned during training for the logitNormal sampling method of \cref{eq.vanilla}, $W$ is initialized with using an uniform law. 

\subsection{Additional Samples:} \label{sup.samples}

\begin{figure}[h]
    \centering
    \includegraphics[width=1\textwidth, trim=0cm 6cm 1.7cm 3cm,clip]{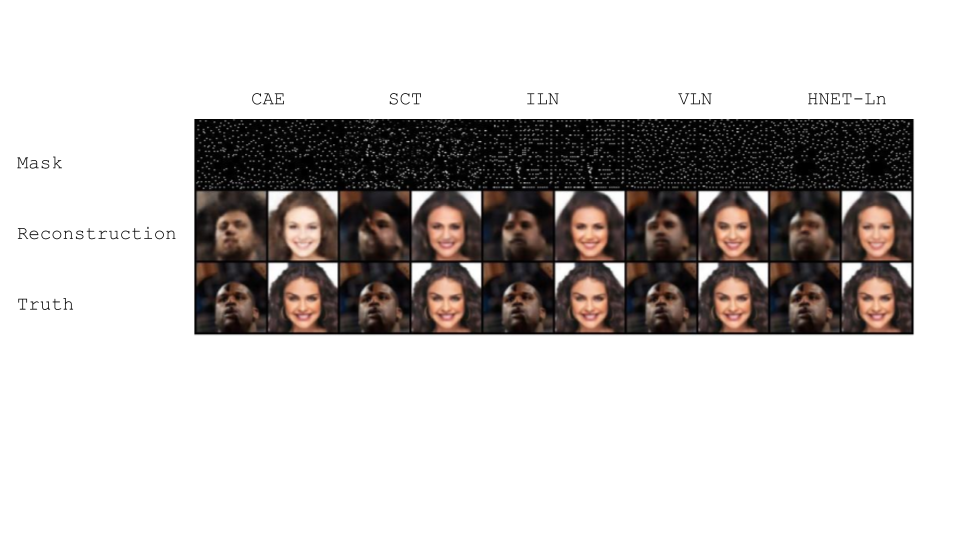}
    \caption{Sample of masks (first row), Reconstruction (second row) and True Data (Last row) for CelebA dataset on all considered algorithms for $200$ features with $\ell_2$-encoding}
    \label{fig:sample_celeba}
\end{figure}

\begin{figure}[h]
    \centering
    \includegraphics[width=1\textwidth, trim=0cm 6cm 1.7cm 3cm,clip]{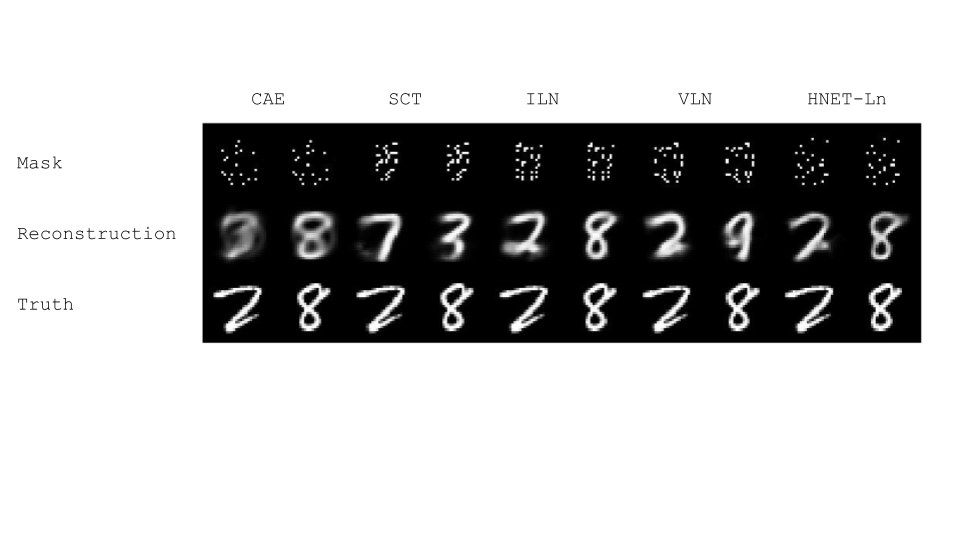}
    \caption{Sample of masks (first row), Reconstruction (second row) and True Data (Last row) for Mnist dataset on all considered algorithms for 20 selected features with $\ell_2$-encoding}
    \label{fig:sample_mnist}
\end{figure}

\begin{figure}[h!]
    \centering
    \includegraphics[width=1\textwidth, trim=0cm 6cm 1.7cm 3cm,clip]{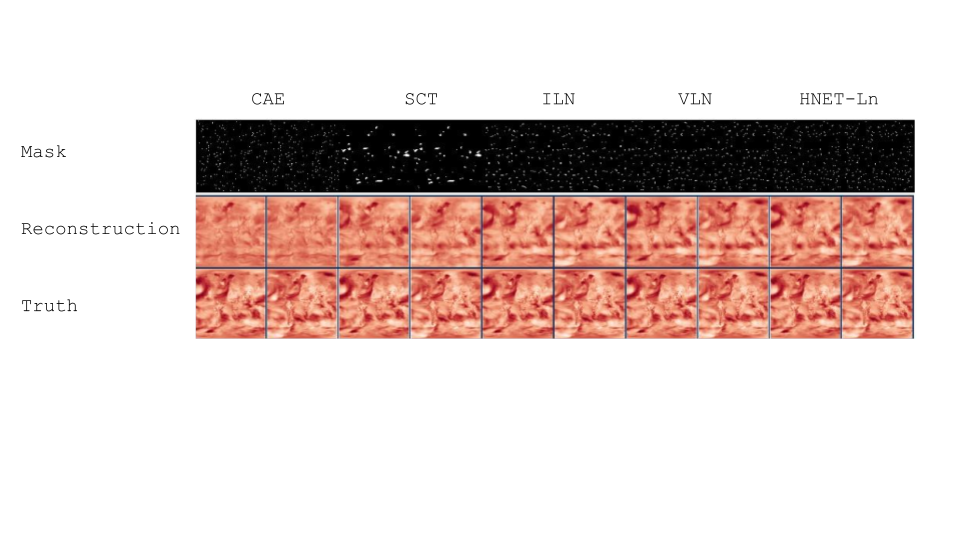}
    \caption{Sample of masks (first row), Reconstruction (second row) and True Data (Last row) for the Geophysical Dataset on all considered algorithms for $200$ features with $\ell_2$-encoding}
    \label{fig:sample_climate}
\end{figure}

\subsection{cGAN Details and Samples} \label{sec.cgan}

\begin{figure}
    \centering
    \includegraphics[width=0.95\textwidth, trim=0.3cm 5.5cm 0.3cm 3.cm, clip]{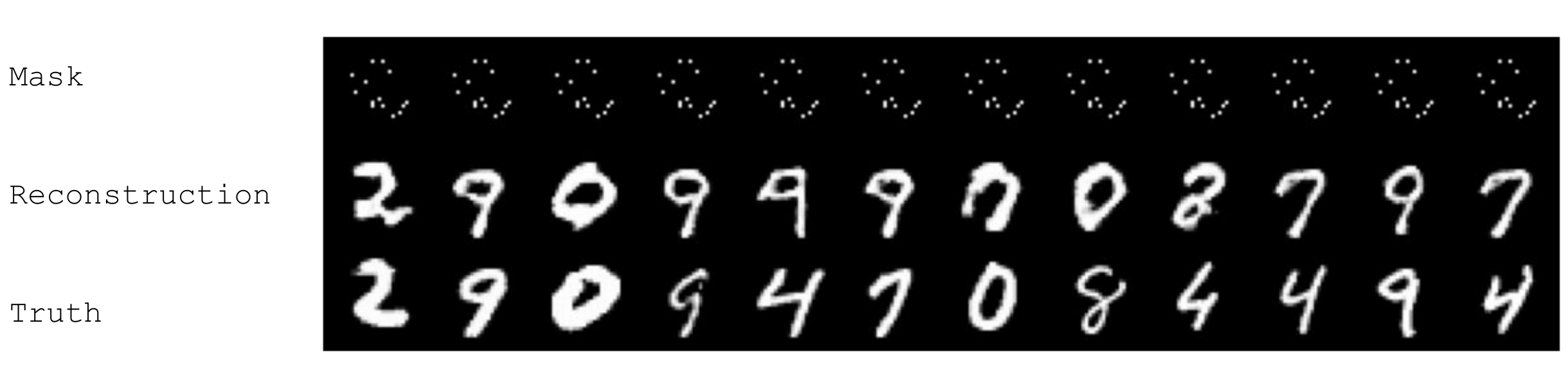}
    \caption{Samples of masks (first row), reconstruction (second row) and true data (last row) for Mnist dataset obtained using a cGAN approach following \cite{isola_image--image_2016}, i.e including a $\ell_1$+Gan loss as reconstruction objective for approximately 15 sampled pixels ($\lambda_{s}=100$)}
    \label{fig:mnist_cgan}
\end{figure}

\begin{figure}
    \centering
    \includegraphics[width=0.95\textwidth, trim=0.3cm 3cm 0.3cm 3.cm, clip]{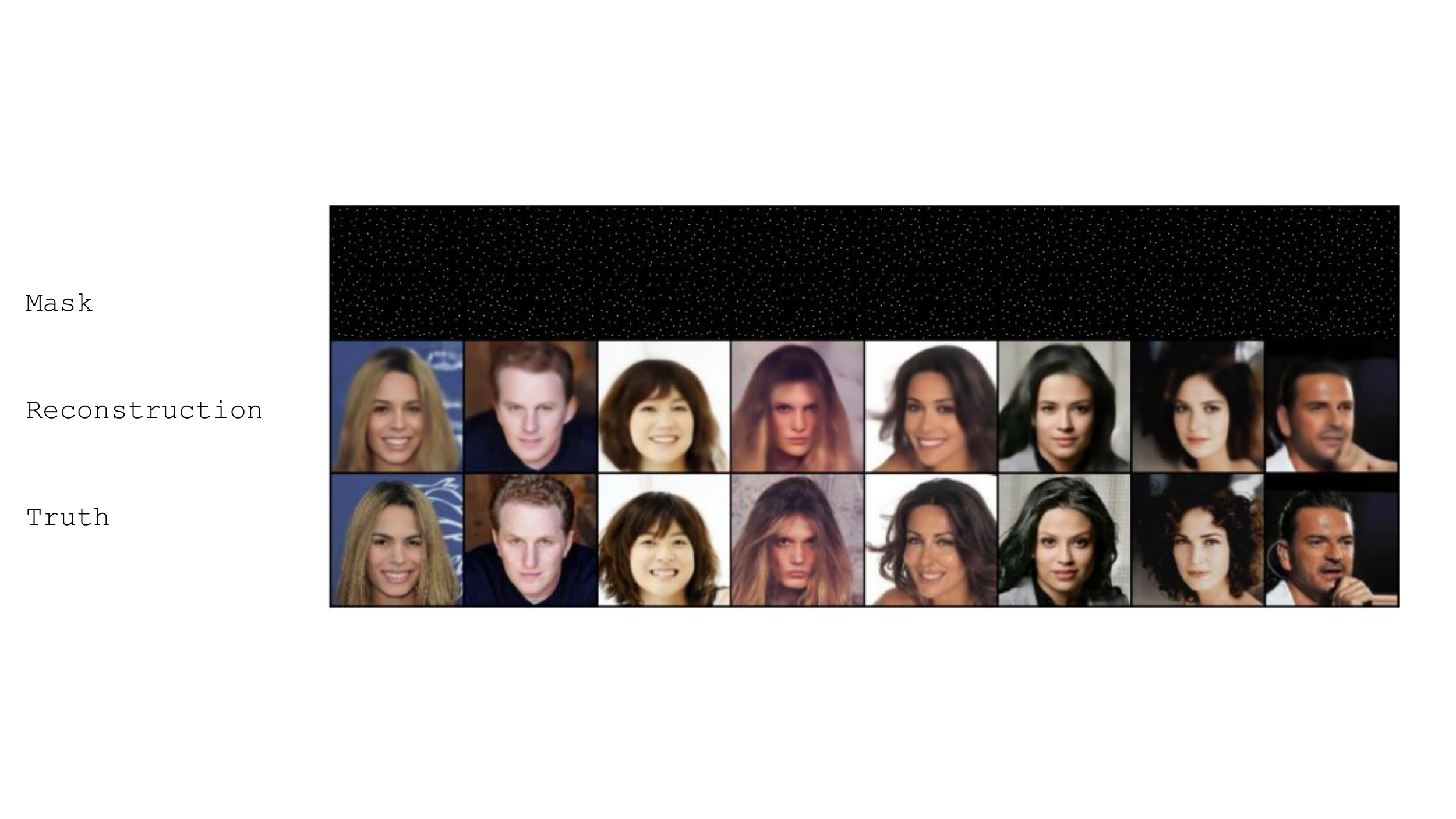}
    \caption{Samples of masks (first row), reconstruction (second row) and true data (last row) for CelebA dataset obtained using a cGAN approach following \cite{isola_image--image_2016}, i.e including a $\ell_1$+Gan loss as reconstruction objective for approximately $1.7\%$ sampled pixels ($\lambda_{s}=100$)}
    \label{fig:celeba_gan}
\end{figure}

Simply speaking, a cGAN has two main learnable functions: a discriminator network with parameters $\psi$ named $D_\psi$ trained to differentiate "true" data labeled as 1 from data generated by $G_\phi$ labeled as 0.
A generative network with parameter $\phi$ denoted $G_\phi$. $G_\phi: \mathbb{R}^p \times \mathbb{R}^{n\times n} \rightarrow \mathbb{R}^n$ takes as input a random variable $\gamma \in \mathbb{R}^p$ and our conditional information $x^{obs}=\Bar{S}_\theta(z) \odot x \in \mathbb{R}^{n\times n}$, and aims at fooling $D_\psi$, making it classify the conditionally generated images as true.
For our experiments we used the cGAN implementation of \cite{isola_image--image_2016} optimizing the following loss, with $x^{obs}=x\odot \Bar{S}_\theta(z)$:
\begin{align}
\min_{\phi, \theta}&\max_{\psi}\mathbb{E}_{z, x} \log D_\psi\big(x,x^{obs}\big) + \ \mathbb{E}_{z,x} \log \big\{ 1 - D_\psi\big( G_\phi(x^{obs}),x^{obs}\big) \big\} \nonumber \\
& + \ \lambda_{sparse} \times \ell_0(\Bar{S}_\theta(z)) + \ \lambda_{rec} \times \ell_1(x - G_\phi(x^{obs})),
\end{align}

Consider $S_\theta$ fixed, one interesting advantage about the cGAN approach is that we can prove that the optimal distribution $p_{G_\phi}$ for $G_\phi$ is given $x^{obs}$: $p_{G_\phi}(x, x^{obs}) = p_{x\sim data}(x|x^{obs})$ which means that $G_\phi$ will sample according to the observed data distribution.

\begin{proof}

\begin{comment}
\begin{align}
\psi^*, \phi^*, S^* = &\argmin_{\phi, S}\max_{\psi}\mathbb{E}_{\gamma, x, x_S} \log D_\psi(x,x_S) \ + \nonumber \\
& \mathbb{E}_{\gamma,x^{S}} \log 1 - D_\psi(G_\phi(\gamma, x_S), x_S) \nonumber \\
& \text{s.t } \ell_0(S_\theta) \leq \lambda 
\end{align}
\end{comment}
To lighten notation, we will use the notation $y=x\odot \Bar{S}_\theta(z)$ as conditioning variable, giving the following game value function:
\begin{align*}
    V(G, D)= \mathbb{E}_{x,y} \log D(x,y) + \mathbb{E}_{z,y} \log \{1 - D(G(z, y), y)\}
\end{align*}
Following \cite{goodfellow_generative_2014}, we can write:
\begin{align*}
    V = &\int_{x, y} \log D(x,y) p_x(x, y)dxdy + \int_{z, y} \log \{1 - D(G(z, y), y) \} p_z(z) p_y(y)dz dy\\
    & \text{if G induce a distribution $p_g$, }\\
    V =& \int_{x, y} [\log D(x, y) p_x(x|y) p_y(y) + \log\{1 - D(x, y)\} p_g(x|y) p_y(y) dx dy] \\
     =& \int_y \Big( \int_x\log D(x, y) p_x(x|y) + \log\{1 - D(x, y)\} p_g(x|y) \Big) p_y(y)dy
\end{align*}
Then classically the maximal value of $x \rightarrow a \log(x) + b \log(1-x)$ is reached in $\frac{a}{a+b}$
Thus, given $y$, the optimal distribution followed by $D$:
$$p_D(x,y) = \frac{p_x(x|y)}{p_x(x|y) + p_G(x|y)}$$
The optimal distribution of $G$ is completely doable at $y$ fixed following the original reasoning of \cite{goodfellow_generative_2014}
\end{proof}

\end{document}